\documentclass[pdflatex,sn-mathphys-num]{sn-jnl}

\usepackage{xcolor}
\usepackage{graphicx}%
\usepackage{multirow}%
\usepackage{amsmath,amssymb,amsfonts}%
\usepackage{amsthm}%
\usepackage{mathrsfs}%
\usepackage[title]{appendix}%
\usepackage{xcolor}%
\usepackage{textcomp}%
\usepackage{manyfoot}%
\usepackage{siunitx}
\usepackage{booktabs}%
\usepackage{algorithm}%
\usepackage{algorithmicx}%
\usepackage{algpseudocode}%
\usepackage{listings}%
\usepackage{subcaption} 
\usepackage{bm}
\usepackage[capitalise]{cleveref}
\usepackage[utf8]{inputenc}
\usepackage{lineno}
\usepackage[font=small,labelfont=bf,figurename=Fig.,labelsep=space]{caption}

\begin{document}

\title[Article Title]{Electromagnetic Navigation for Femoral Osteotomy Using High-Accuracy X-ray-to-CT Registration}

\author*[1,2]{\fnm{Roman} \sur{Flepp\protect\textsuperscript{$\dagger$}}}\email{roman.flepp@kispi.uzh.ch}

\author*[1,3]{\fnm{Arend} \sur{Nieuwland\protect\textsuperscript{$\dagger$}}}\email{arend.nieuwland@balgrist.ch}

\author[2]{\fnm{Bastian} \sur{Sigrist}}
\author[2]{\fnm{Philipp} \sur{Fürnstahl}}
\author[2]{\fnm{Lilian} \sur{Calvet\protect\textsuperscript{$\ddagger$}}}
\author[1,3]{\fnm{Thomas} \sur{Dreher\protect\textsuperscript{$\ddagger$}}}

\affil[1]{\orgname{Department of Pediatric Orthopedics and Traumatology, University Children's Hospital Zürich}, \orgaddress{\country{Switzerland}}}
\affil[2]{\orgdiv{Research in Orthopedic Computer Science}, \orgname{University Hospital Balgrist, University of Zurich}, \orgaddress{\country{Switzerland}}}
\affil[3]{\orgdiv{Department of Orthopedic Surgery}, \orgname{University Hospital Balgrist, University of Zurich}, \orgaddress{\country{Switzerland}}}
\affil[\protect\textsuperscript{$\dagger$}]{These authors contributed equally}
\affil[\protect\textsuperscript{$\ddagger$}]{These authors jointly supervised this work and share last authorship}

\abstract{

\textbf{Purpose:} Accurate execution of preoperative plans in corrective femoral osteotomies remains a substantial challenge. Current techniques are limited by variable accuracy, invasiveness, and substantial radiation exposure, with free-hand methods and patient-specific instrumentation (PSI) often requiring \num{>30} and \num{>6} fluoroscopic images, respectively. This work addresses the need for an accurate navigation system that minimizes dissection and intraoperative fluoroscopy.

\noindent\textbf{Methods:} We present an integrated, electromagnetic tracking (EMT)-based navigation system for femoral osteotomies. The system couples CT-based preoperative bone planning with one time intraoperative C-arm calibration and accurate X-ray-to-CT registration, based on two fluoroscopic images acquired at initialization. This registration is followed by real-time, fluoroscopy-free EMT-based navigation of the saw blade, and subsequently of the bone fragments, with respect to the preoperative plan, and is compatible with uniplanar and biplanar osteotomies.

\noindent\textbf{Results:} In a feasibility study utilizing 18 synthetic femora, EMT guidance significantly outperformed free-hand execution for total angular error ($(3.05 \pm 0.75)^\circ$ vs.\ $(6.32 \pm 2.36)^\circ$, $p=0.031$). No EMT-guided trials exceeded the \num{>5}° clinical threshold, whereas free-hand resulted in 4 outliers of 6 trials while assuming the same minimal surgical exposure for both techniques. The system achieved statistical equivalence ($\pm 2^\circ$, $\pm 2\,\text{mm}$) to PSI for total angular ($p \le 0.02$) and total translational ($p=0.048$) errors with no significant differences in user questionnaire scores.

\noindent\textbf{Conclusion:} The proposed EMT-based system enables transfer of preoperative plans to the OR using two fluoroscopic images and provides real-time intraoperative guidance. In a synthetic-bone feasibility study, it substantially improved correction accuracy over free-hand execution and achieved equivalent accuracy to PSI, whilst no additional surgical exposure is needed and requiring substantially fewer fluoroscopic images, motivating subsequent cadaveric and clinical validation.
}


\maketitle
\section{Introduction}

Corrective femoral osteotomies are an established treatment for extra-articular deformities and torsional malalignment of the femur in pediatric and adult patients \cite{dreher2012long}. By restoring joint congruency and the mechanical axis of the lower limb, these procedures can prevent gait disturbances, pain, and joint damage \cite{dreher2012long, buly2018femoral}. However, achieving the planned correction is critical: residual malalignment exceeding $5^\circ$ has been identified as a predictor for osteoarthritis progression and the need for subsequent corrective procedures \cite{sharma2001role}. Consequently, technology-assisted execution must not only improve mean accuracy but also reliably reduce outliers and maintain postoperative alignment errors below the $5^\circ$ threshold.

While modern planning workflows rely on 3D imaging (e.g., CT) to quantify deformities and simulate osteotomies, transferring these plans from preoperative planning to intraoperative execution remains a challenge \cite{Neopoulos2025}. Conventional free-hand execution depends heavily on surgeon experience and iterative fluoroscopic imaging to cognitively reconstruct the 3D alignment from 2D views and incrementally adjust the correction. This often leads to inaccurate corrections and substantial intraoperative radiation exposure \cite{Shi2019, kriechling2025radiation, grillo20253d}. In a 33-patient clinical study, Shi et al.\ reported that free-hand distal femoral osteotomies required an average of 34.7 fluoroscopic images \cite{Shi2019}.

To address these limitations, PSI has been introduced by constraining tool placement to the preoperative plan. This technique utilizes custom 3D-printed guides, designed by specialists to fit the patient's bone surface and mechanically direct surgical tools to execute the planned corrections \cite{Jud_2020,Neopoulos2025}. Although PSI can reduce fluoroscopy usage (6.1 images \cite{Shi2019}), it introduces logistical burdens, manufacturing lead times and additional costs, while reducing intraoperative flexibility and being a demanding technique \cite{khela2025psi_review, grillo20253d}. Furthermore, PSI commonly requires bony structure exposure and larger incisions for guide placement \cite{khela2025psi_review, Neopoulos2025}. Such extensive exposures are particularly undesirable in pediatric populations, where minimally invasive approaches are preferred.

Computer-assisted navigation has increasingly become a valuable alternative. Optical tracking systems can provide highly accurate real-time feedback, but they rely on an unobstructed line-of-sight and often necessitate bulky bone-mounted reference markers to ensure accuracy, as direct visual skin or limb tracking is less accurate and susceptible to soft-tissue movements \cite{casari2021ar, song2016computer, Sorriento2020}. Nonetheless, optical tracking systems have demonstrated good performance for certain types of surgeries e.g. cranial and spinal \cite{stryker_q_guidance}. EMT serves as a compelling alternative, enabling accurate six-degree-of-freedom (6-DoF) tracking of miniature sensors without line-of-sight requirements \cite{Sorriento2020}. EMT has been commercially used for spinal surgeries, however prior applications of EMT in femoral osteotomies have largely treated it as a measurement or verification tool rather than a navigation system \cite{MedtronicStealthS8, geisbusch2025multiplanar}.

Despite these advantages, the adoption of EMT for femoral osteotomies is hindered by the challenge of CT-to-EMT registration. To navigate, the preoperative CT plan must be accurately registered to the physical position of the EMT sensors, which requires an accurate CT-to-EMT registration. State-of-the-art registration methods typically rely on paired-point matching followed by surface matching \cite{geisbusch2025multiplanar, liu_2020, MedtronicStealthS8} or fluoroscopic intensity matching \cite{Gopalakrishnan_2024, MedtronicStealthS8}. Paired-point matching requires the surgeon to manually palpate anatomical landmarks, which is error-prone and time-consuming. Surface matching requires digitizing large areas of the bone surface, necessitating extensive soft-tissue stripping \cite{geisbusch2025multiplanar, liu_2020}. While fluoroscopic intensity-based registration offers a non-invasive alternative, existing solutions often fail to achieve necessary clinical robustness \cite{Flepp2025} or lack automation \cite{imfusion_software}.

To address these limitations, we propose to use a recently published, highly automated and accurate X-ray-to-CT registration method \cite{Flepp2025} and integrating it with EMT into a unified, low-radiation navigation system. The proposed workflow unifies three core components: (i) preoperative CT-based planning, (ii) an automatic two-view C-arm calibration and X-ray-to-CT registration pipeline, and (iii) real-time intraoperative electromagnetic tracking. These elements are integrated into a custom graphical user interface (GUI) that supports uniplanar and biplanar osteotomies through accurate saw blade and bone fragment tracking.

The contributions of this paper are:
\begin{enumerate}
    \item[(i)] A practical, automated femoral osteotomy navigation system integrating two-view fluoroscopy calibration, automatic X-ray-to-CT registration, and EMT;
    \item[(ii)] A system-level validation of EMT interference and calibration; and
    \item[(iii)] A feasibility study on synthetic bone models (two surgeons, three osteotomy scenarios) benchmarking our system against free-hand and PSI. This preliminary study demonstrates a significant reduction in angular error relative to free-hand techniques and statistical equivalence to PSI was demonstrated.
\end{enumerate}

\begin{figure}[h]
    \centering
    \includegraphics[width=0.99\linewidth]{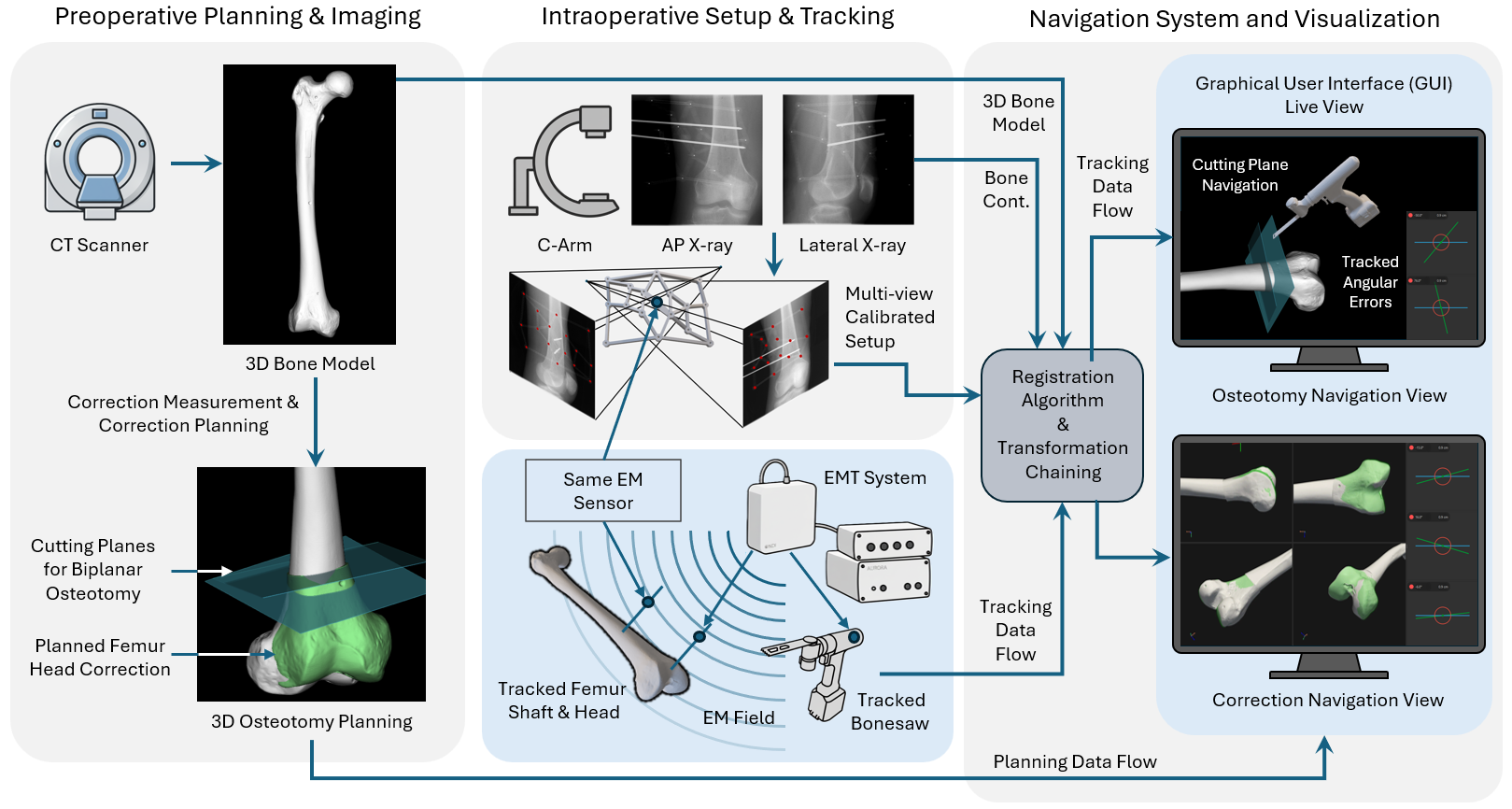}
    \caption{Overview of our EMT-based system, from preoperative planning (left), intraoperative calibration, X-ray-to-CT registration, EMT (middle), visualizations and dataflows (right).}
    \label{fig:overview}
\end{figure}

\section{Methodology}
\label{sec:method}
Our proposed navigation workflow synergizes the fully automated X-ray-to-CT registration framework recently proposed by Flepp et al.~\cite{Flepp2025} with the intraoperative advantages of EMT. While Flepp et al. provide a robust solution for static one-time registration and C-arm calibration, we propose to extend this capability to dynamic surgical guidance.

The protocol operates in two distinct phases: \textit{initial registration} and \textit{real-time navigation}, while maintaining a minimal surgical footprint, utilizing only the two Kirschner wires (K-wires), consistent with the standard free-hand technique. The initial registration establishes the spatial link between the preoperative CT bone model and the EMT coordinate system. This is achieved using a custom C-arm calibration phantom of known geometry, which is rigidly co-mounted with an EMT sensor onto a K-wire fixed to the femur. The system derives the accurate pose of the bone relative to the EMT sensors by computing a transformation chain linking the sensors, phantom, fluoroscopic images, and CT data. Once registered, the phantom is detached, leaving the sensor on the K-wire to provide real-time guidance of the bone fragments.

\subsection{Hardware}
While our proposed navigation workflow is adaptable to other setups, this study utilized the commercially available Aurora EMT system (Northern Digital Inc., Canada). This system provides a cubic workspace of edge length $50\,\text{cm}$ and sub-millimetre accuracy using 6-DoF sensors, covering the distal femoral workspace. The sensors are attached to the relevant objects like the Colibri II (DePuy Synthes, Switzerland)and $3\,\text{mm}$ K-wires (MEDE Technik, Germany) using custom 3D-printed fixtures, engineered for vibration resistance and intraoperative motions. Intraoperative fluoroscopic images were acquired using the Cios Spin (Siemens, Germany) C-arm system and final bone fragment fixation was done using the pediatric LCP plate system (DePuy Synthes, Switzerland).
\newline

\begin{figure}[h]
    \includegraphics[width=\linewidth]{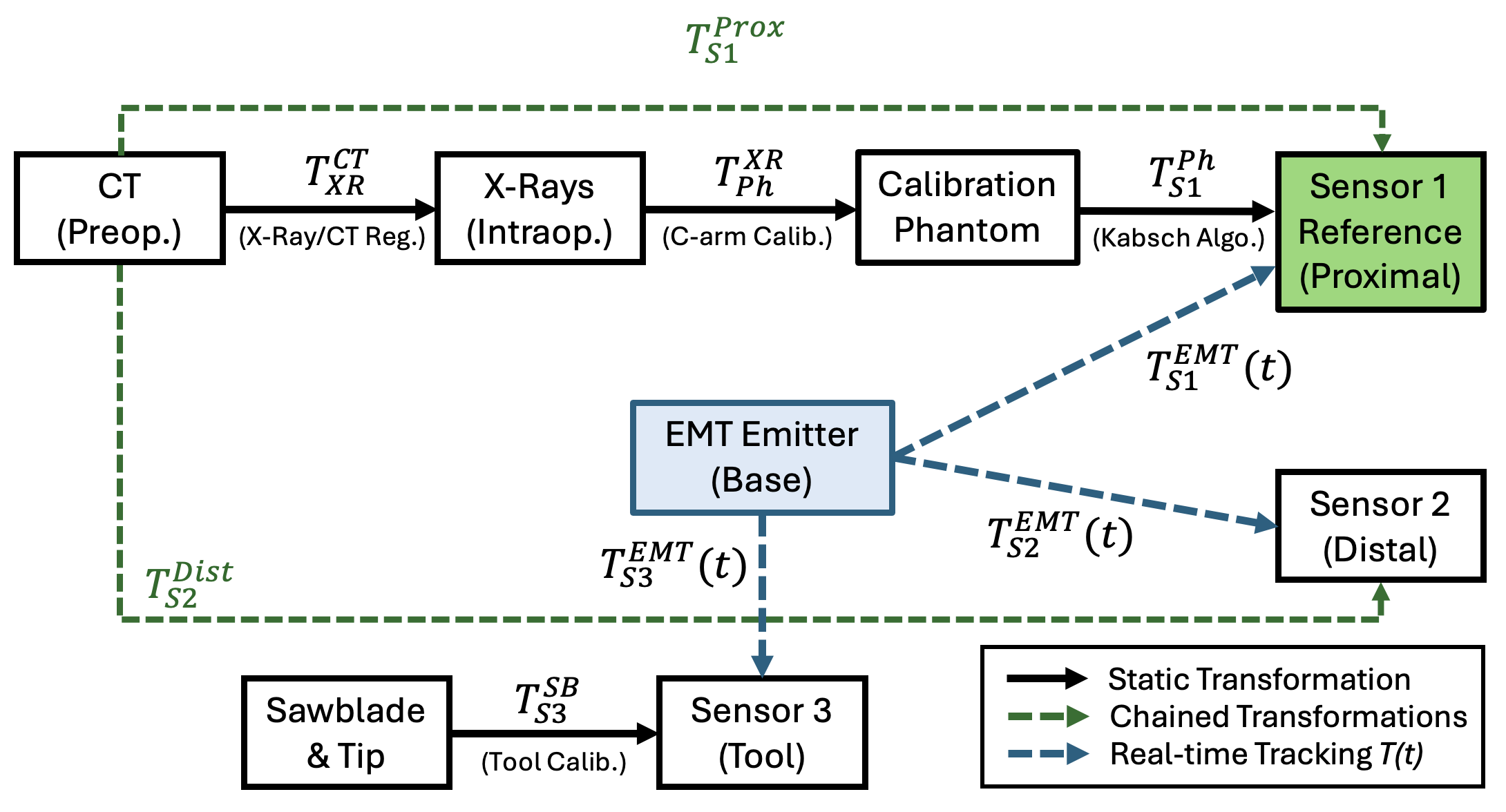}
    \caption{Coordinate frames and transformations. The transformation $\mathbf{T}_A^B$ maps a point from coordinate frame $A$ to frame $B$.}
    \label{fig:frames_vis}
\end{figure}

\subsection{Software and Tracking}
Our primary contribution is the navigation software, which fuses X-ray-to-CT registration and signals from the EMT sensors with the preoperative CT plan. The software manages the calibration of X-ray images, registers them to the preoperative CT volume, visualizes the osteotomy plan, processes real-time sensor transformations, and enables surgical navigation as seen in \cref{fig:overview} and \cref{fig:frames_vis}. We assume that the preoperative CT scan and surgical plan are available prior to this workflow.
\newline

\begin{figure}[h]
    \includegraphics[width=\linewidth]{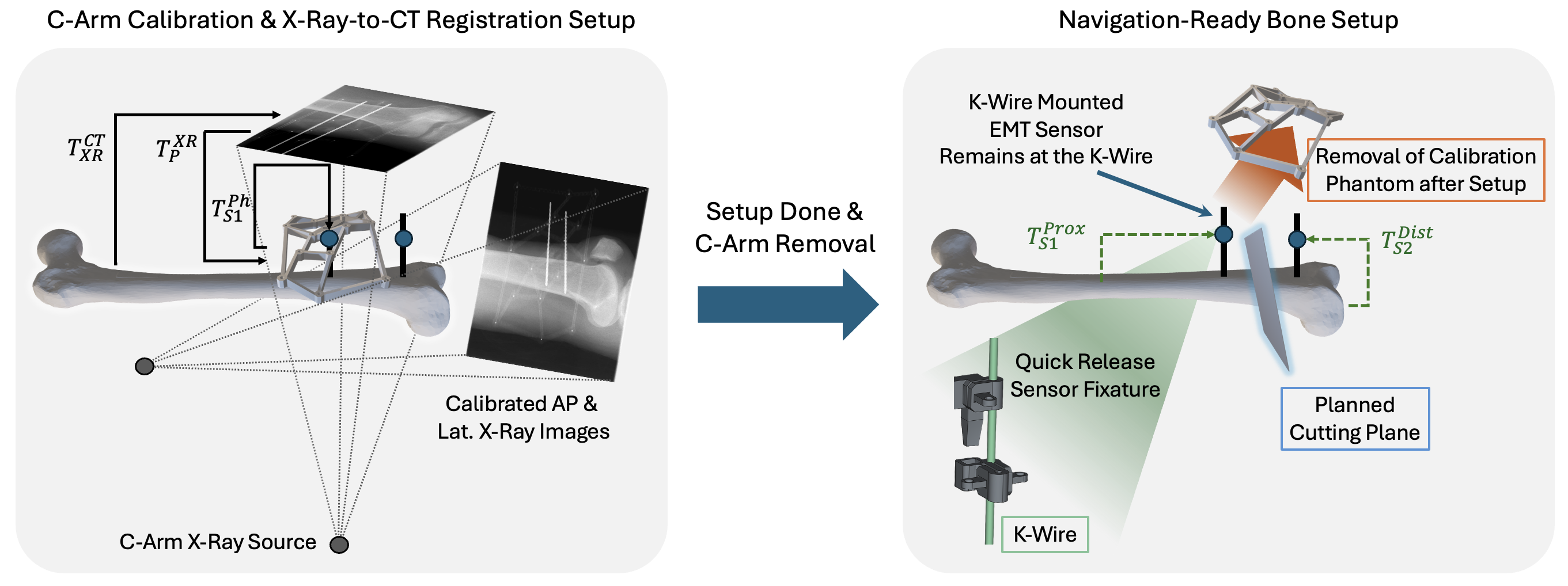}
    \caption{A detailed view of the middle panel from \cref{fig:overview}. Left: The calibration phantom and the K-wire fixed sensors. Right: After the X-ray images calibration and X-ray-to-CT registration, the phantom is removed. The navigation-ready setup is shown with a detailed view of the quick release mechanism of the calibration phantom, with the upper part fixated to the phantom and the lower part containing \textit{sensor 1}.}
    \label{fig:pnpf_phantom}
\end{figure}

\noindent \textbf{Preoperative phantom \& sleeve calibration} The preoperative system setup begins with the registration of the C-arm calibration phantom to the EMT system yielding $\mathbf{T}_{\text{S1}}^{\text{Ph}}$. The phantom as seen in \cref{fig:pnpf_phantom}, comprises 16 radiopaque stainless-steel beads arranged in a well-defined geometric configuration, ensuring high fluoroscopic visibility. An EMT sensor is attached to the phantom via a custom release mechanism which ensures that the sensor does not move relative to the registered bone when the phantom is being removed from the K-wire, namely right after the X-ray-to-CT registration. It consists of a slot where the sensor is attached rigidly, preventing any residual movement. After attachment, the bead positions are digitized using the tracked NDI Aurora 6-DoF Probe. This probe functions like a stylus, providing measurements of its tip's position in the electromagnetic field. The Kabsch algorithm is employed to register the digitized points to the virtual model, yielding $\mathbf{T}_{\text{Ph}}^{\text{S1}}$. Consequently, \textit{sensor 1} serves as the global reference frame. This preoperative step is repeated analogously for a sleeve to perform sawblade calibration, utilizing predefined marker holes sampled by the probe to derive $\mathbf{T}_{\text{SB}}^{\text{S3}}$.
\newline

\noindent \textbf{Intraoperative calibration and registration framework} To establish the coordinate transformations of X-ray images to CT $\mathbf{T}_{\text{XR}}^{\text{CT}}$ and phantom to X-ray images $\mathbf{T}_{\text{Ph}}^{\text{XR}}$ as in \cref{fig:frames_vis}, we employ a well-known workflow for X-ray calibration and recent X-ray-to-CT registration method, as detailed by Flepp et al.~\cite{Flepp2025}.

First, two K-wires are intraoperatively inserted into the bone, whose locations adhere to standard free-hand workflow. The calibration phantom, equipped with \textit{sensor 1}, is rigidly fixed to the proximal wire, while a \textit{sensor 2} is mounted to the distal wire. The C-arm is positioned to acquire sequential anterior-posterior (AP) and lateral views, ensuring that the metallic beads remain within the field of view, and subsequently moved away again. As reported in \cite{Flepp2025} a U-Net extracts the bead centroids which are then used to find the intrinsic and extrinsic parameters of the C-arm, yielding $\mathbf{T}_{\text{Ph}}^{\text{XR}}$.
To resolve the X-ray-to-CT registration transformation $\mathbf{T}_{\text{XR}}^{\text{CT}}$ from the calibrated views, we employ the substructure contour-based optimization method \cite{Flepp2025}. Although our system is registration method agnostic, this specific algorithm provides high robustness and accuracy while not requiring initialization typically used in intensity-based approaches, which is often performed manually \cite{imfusion_software}. Concretely, the optimization consists of an iterative closest point (ICP) scheme which aligns reprojected preoperative CT bone contours to the intraoperatively segmented X-ray image bone contours. In the distal femur use-case, this technique demonstrated a mean reprojection distance (mRPD) of $0.67\,\text{mm}$~\cite{Flepp2025}.
Contrary to the use of physical guides in PSI, this contour-based method allows to assess the quality of the alignment between the CT bone model and the intraoperative X-ray images, after the X-ray-to-CT registration, through visual inspection of the contour matching and also through palpation of known anatomical structures with navigated tools. Finally, the calibration phantom is detached, retaining the wire-fixed \textit{sensor 1} as reference frame. Once detached, this setup maintains the ergonomic flexibility of the standard free-hand workflow.
\newline

\begin{figure}[h]
    \includegraphics[width=\linewidth]{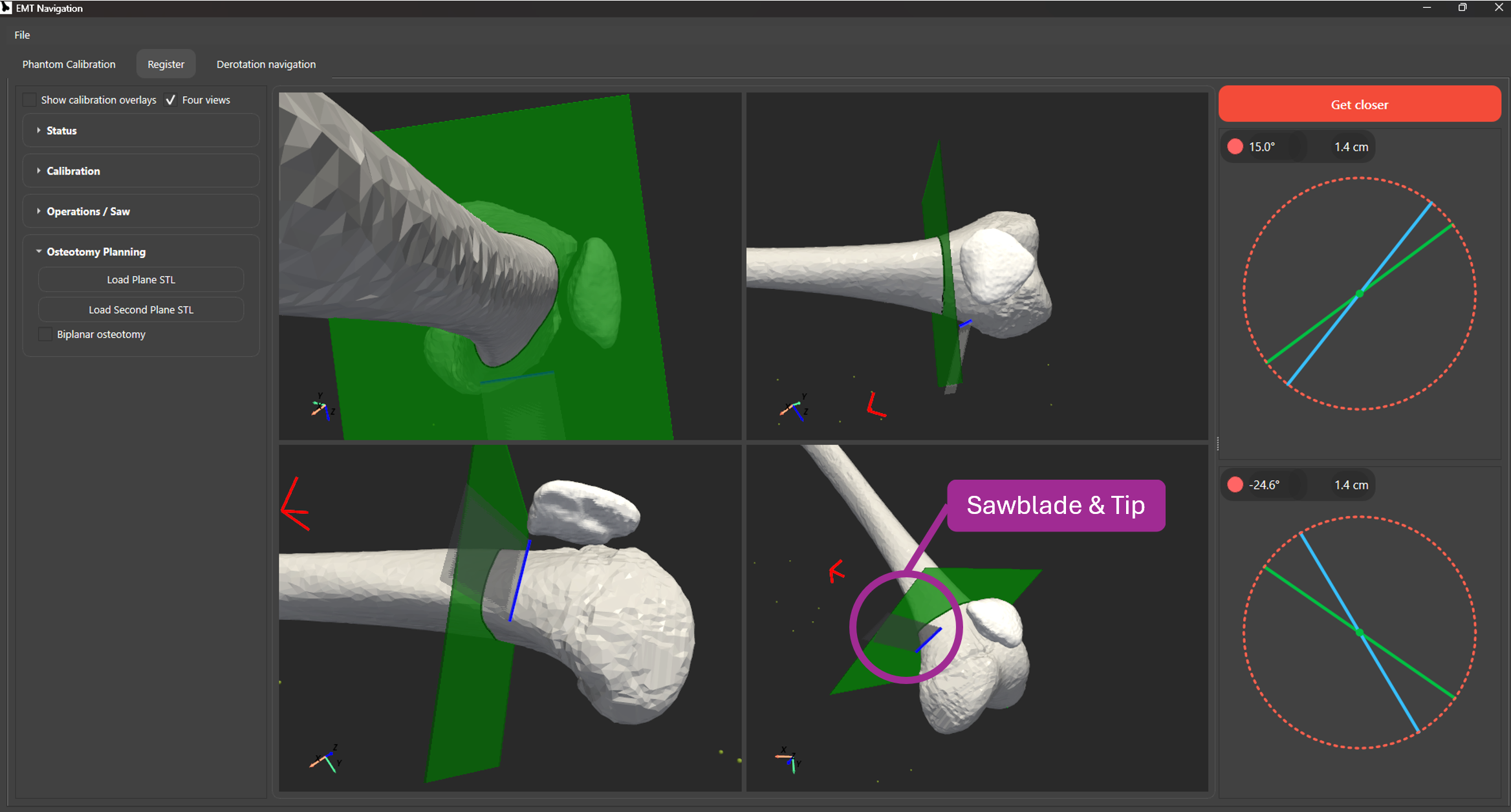}
    \caption{The osteotomy navigation view in the GUI, showing the planned cutting plane in transparent green, the current position of the bone saw tip in blue and the sawblade in transparent grey. The levers on the right side show the angular errors of the saw plane relative to the cutting plane and the red circle the distance metrics. On the top right is a status bar, improving the usability.}
    \label{fig:GUI_1}
\end{figure}

\noindent\textbf{Osteotomy cut navigation} 
To provide the surgeon with cut alignment of the osteotomy, the system must relate the pose of the sawblade within the CT coordinate system. This is achieved by chaining the static registration results with the dynamic measurements $\mathbf{T}_{SB}^{CT}(t) = \mathbf{T}_{XR}^{CT} \cdot \mathbf{T}_{Ph}^{XR} \cdot \mathbf{T}_{S1}^{Ph} \cdot \left(\mathbf{T}_{S1}^{EMT}(t)\right)^{-1} \cdot \mathbf{T}_{SB}^{EMT}(t)$. This allows the system to track the position of the saw tip and blade as seen in \cref{fig:GUI_1}. The real-time visualization also serves as a registration monitor. The surgeon can at any point palpate a known anatomical landmark with the tracked tool tip or the 6-DoF probe to verify that the virtual representation remains aligned with the anatomy. Furthermore, it provides depth verification, which is fundamental for clinical safety, minimizing the risk of injury to critical neurovascular structures.
\newline

\noindent\textbf{Correction navigation}
Following the navigated cuts, the proximal and distal femur fragments are mechanically separated. To guide the subsequent corrections, the system must track the relative pose of the distal fragment with respect to the proximal part in real-time. In contrast to the sawing phase, the femur can no longer be treated as a single rigid body; therefore, using the dynamic transformations $\mathbf{T}_{\text{EMT}}^{\text{S1}}(t)$ and $\mathbf{T}_{\text{EMT}}^{\text{S2}}(t)$ the relative transformation $\mathbf{T}_{\text{Prox}}^{\text{Dist}}(t)$ is computed as follows: $\mathbf{T}_{\text{Prox}}^{\text{Dist}}(t) = \mathbf{T}_{\text{S2}}^{\text{Dist}} \cdot \left(\mathbf{T}_{\text{S2}}^{\text{EMT}}(t)\right)^{-1} \cdot \mathbf{T}_{\text{S1}}^{\text{EMT}}(t) \cdot \mathbf{T}_{\text{Prox}}^{\text{S1}}$.

\begin{figure}[h]
    \includegraphics[width=\linewidth]{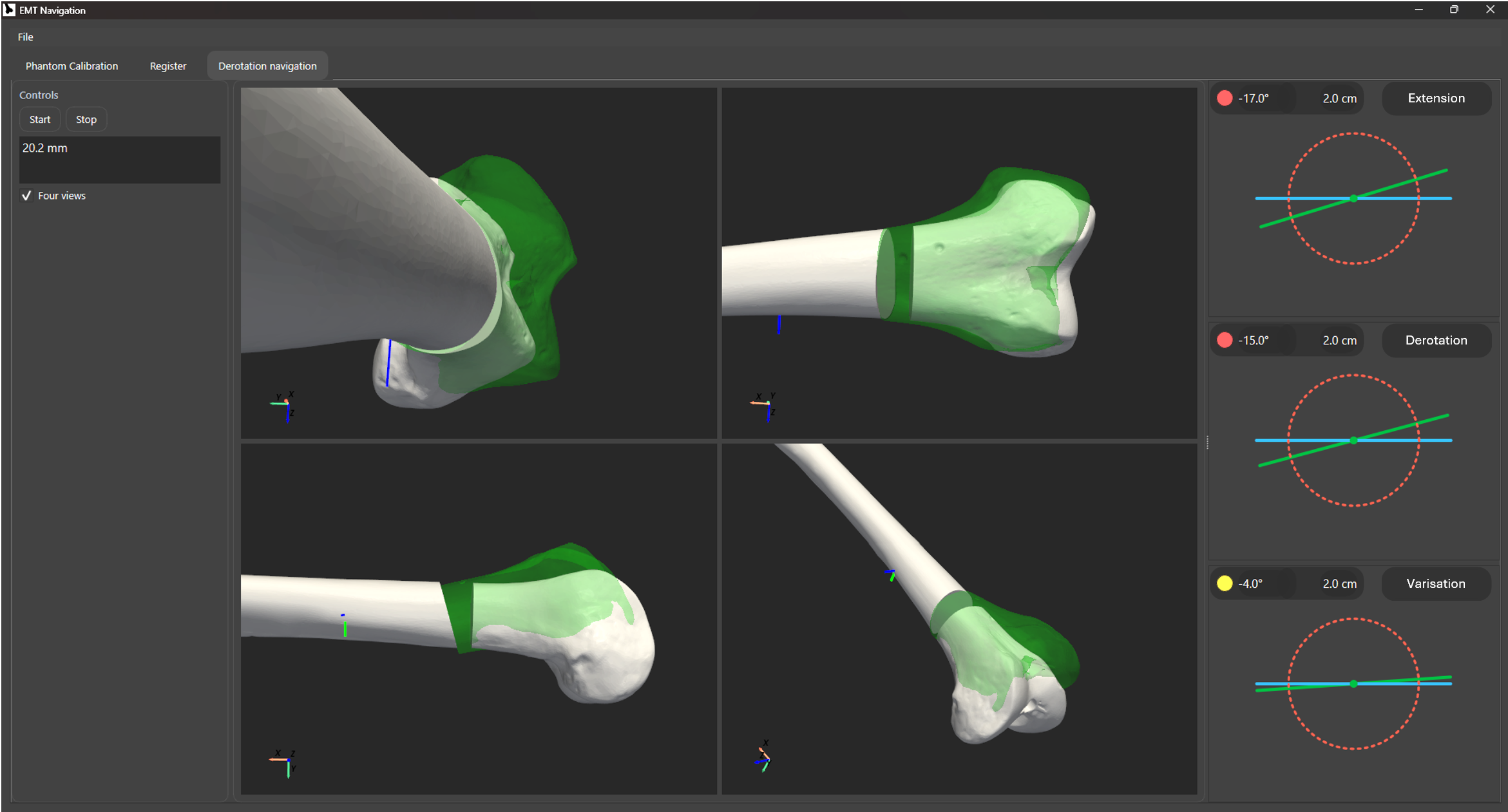}
    \caption{The correction navigation view in the GUI, showing the planned target position in transparent green and the current position in white. The levers on the side show the angular errors and the circle the distance metrics.}
    \label{fig:GUI_2}
\end{figure}

This approach enables real-time visualization of the relative fragment movements, facilitating leg manipulation during the correction. The system continuously compares the live alignment against the planned target, decomposing deviations into clinically relevant translational and rotational components (derotation, extension, varisation) displayed on the GUI (\cref{fig:GUI_2}).

\section{Experiments}
\label{sec:xp}

We validated the proposed EMT workflow through interference characterization and a synthetic femoral bone feasibility study. The primary objective of the study was to determine if EMT guidance improves osteotomy accuracy compared to free-hand execution and if comparable accuracy to PSI is achieved.

\subsection{System Characterization and Technical Validation}
\label{sec:tech_val}

\noindent\textbf{EMT interference and calibration}
The performance of the navigation system relies on the intrinsic accuracy of the EMT system and our calibration steps. While the manufacturer specifies an accuracy of $0.48\,\text{mm}$ and $0.30^\circ$ under idealized conditions, ferromagnetic surgical instruments or running motors can distort the magnetic field. Consequently, we characterized the interference of the Synthes Colibri II bone saw relative to a stationary sensor. Measurements were acquired at varying proximities ($1\,\text{mm}$ to $20\,\text{mm}$) under two conditions: \textit{idle} (motor unpowered) to simulate trajectory alignment, and \textit{active} (motor engaged) simulating the cutting phase. Additionally, the impact of close sensor-to-metal contact was evaluated by attaching sensors directly to standard $3\,\text{mm}$ stainless steel K-wires. Finally the phantom and sleeve calibration method using the Aurora probe was evaluated using the calibrated saw-blade and touching known points in space.

The interferences and the calibration method were quantified as follows:
\begin{itemize}
    \item \textbf{Idle power tool:} Positioning the idle saw within $1$\,mm of a sensor induced a mean angular deviation of $5.1^\circ$ and a mean translational deviation of $8$\,mm across 20 trials. Distortion decayed rapidly with distance, returning to baseline levels (orientation deviations $<0.30^\circ$) at distances $\geq 20$\,mm.
    \item \textbf{Active power tool:} The active saw motor generated electromagnetic noise and vibrations that degraded tracking accuracy beyond the clinical tolerance of $5^\circ$ at $20$\,mm. An intermittent verification workflow is enforced: the cutting plane is aligned with the saw in the idle state, the cut is performed, and the motor is momentarily halted to verify depth and trajectory against the surgical plan.
    \item \textbf{K-wires:} Direct attachment of sensors to $3$\,mm stainless steel K-wires resulted in negligible additional error relative to baseline, validating the proposed bone-mounting strategy.
    \item \textbf{Calibration:} Repeated measurements revealed an accuracy of $0.53$\,mm for the calibrated saw-tip, close to the system's intrinsic accuracy, confirming the validity of the calibration method.
    
\end{itemize}
Based on the idle-state decay profile, we enforced a minimum $20$\,mm safety margin for our sensor fixtures and implemented a software alert if this margin is violated.

\subsection{Study Design}
\label{sec:dataset}
Three femoral osteotomy scenarios of increasing clinical complexity including the PSI guides were planned and designed by the Balgrist center for 3D preoperative planning and 3D printing. These scenarios comprise:

\begin{description}
    \item[Level A:] A standard single-cut osteotomy requiring a pure $15^\circ$ derotation (external rotation) representing baseline complexity.
    \item[Level B:] A complex cut requiring a compound correction of $15^\circ$ derotation (external rotation) and $10^\circ$ varus adjustment.
    \item[Level C:] A biplanar osteotomy including a multi-dimensional correction: $17^\circ$ extension, $15^\circ$ derotation (external rotation), and $3^\circ$ varus simulating a complex deformity correction where free-hand execution is demanding.
\end{description}

\begin{figure}[t!]
    \centering
    \includegraphics[width=0.95\linewidth]{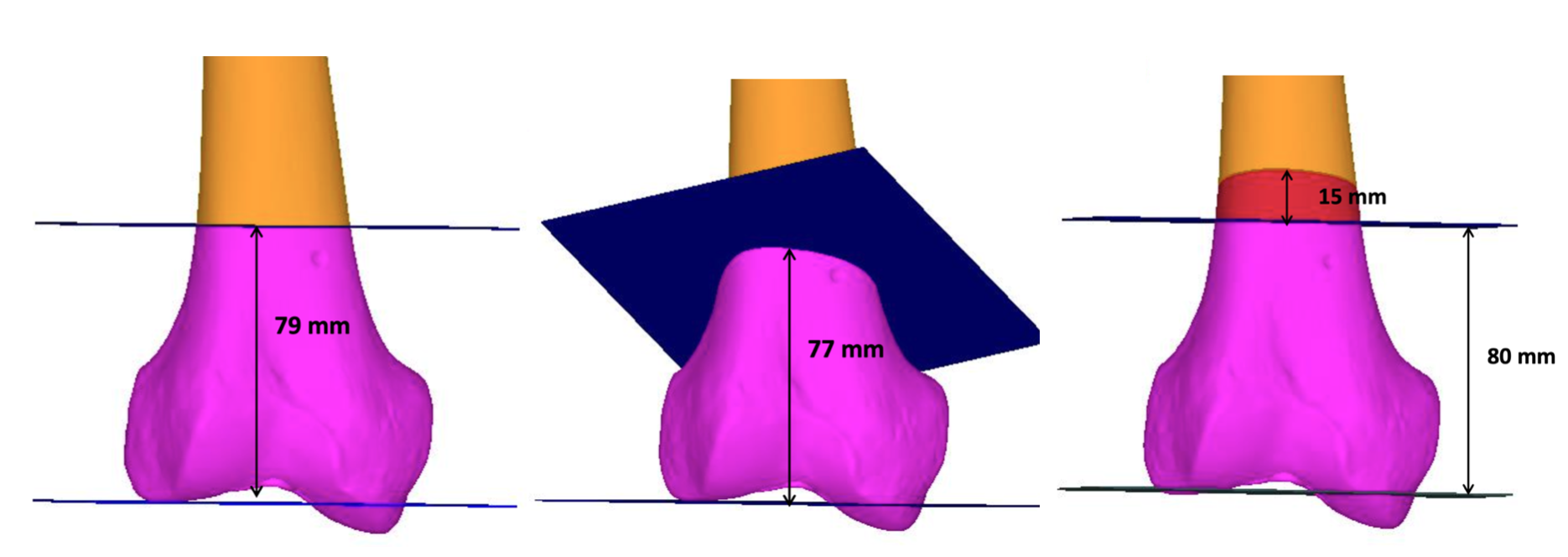}
    \caption{Visualization of the three planned osteotomies used in the study: (A) Simple orthogonal cut, (B) Complex multi-correction cut, and (C) Complex biplanar osteotomy.}
    \label{fig:planned_corrections}
\end{figure}

The osteotomy plans are visualized in \cref{fig:planned_corrections}. The study followed a block design: two senior orthopedic surgeons performed the correction for all three complexity levels using all three navigation methods in multiple sessions separated by at least 2 days, resulting in a total of $N=18$ trials. The surgeons were experienced with free-hand and PSI osteotomies and received a foundational training with the proposed EMT system. All trials were performed on replicas of a single femoral bone model (SYNBONE, Switzerland).

\subsection{Evaluation Metrics}
\label{sec:evalmet}

To quantify surgical outcome, postoperative CT segmentations were compared against the preoperative plan. Consistent with standard computer-assisted orthopedic surgery validation protocols, rotational and translational deviations were calculated to assess joint mechanics. Specifically, the postoperative segmented CT bone mesh was aligned to the preoperative proximal bone mesh using ICP. Subsequently, the planned (preoperatively corrected) distal bone mesh was registered to the corresponding postoperative distal fragment using ICP. The resulting transformation matrix yielded the following error analytics:

\begin{itemize}
    \item \textbf{Angular Accuracy ($E_{\mathrm{rot}}$):} The absolute residual rotation, obtained by decomposing the final pose into clinically interpretable Euler angle components that directly correspond to the planned osteotomy corrections (derotation, varisation, and extension) is our primary clinically relevant metric. Reporting the total euclidean angular error and its components enables a direct assessment of whether the intended angular corrections were achieved.
    \item \textbf{Outlier rate ($E_{\mathrm{rot}}>5^\circ$):} The number of trials with at least one residual angular component exceeding the clinically relevant $5^\circ$ threshold \cite{sharma2001role}. This endpoint directly quantifies whether a method reduces clinically meaningful malalignment outliers.
    \item \textbf{Translational Accuracy ($E_{\mathrm{trans}}$):} The Euclidean norm (in mm) of the residual translation vector of the final pose. This secondary metric captures unintended fragment shifts, which can affect offset, effective limb length, and soft-tissue tension.

\end{itemize}

\noindent\textbf{Statistical analysis} The data points were analyzed as a randomized complete block design with six matched blocks (2 surgeons $\times$ 3 difficulty levels). Although Shapiro-Wilk tests indicated no significant deviation from normality (all $p > 0.05$), non-parametric methods were prioritized due to the limited statistical power inherent in the small sample size per method ($n=6$). Pre-planned contrasts assessed the superiority of EMT vs.\ free-hand and the equivalence of PSI vs.\ EMT. Superiority was evaluated using a one-sided Wilcoxon signed-rank test ($\alpha = 0.05$). Equivalence was assessed using a non-parametric two one-sided tests (TOST) procedure with symmetric bounds of $\delta=2.0^\circ$ and $\delta=2.0\,\text{mm}$. Results report differences with 95\% confidence intervals (CI) for superiority and 90\% CI for equivalence. Descriptive statistics denote pooled mean $\pm$ standard deviation (SD).

\section{Results}
\label{sec:results}

\begin{figure}[t!]
    \centering
    \includegraphics[width=0.99\linewidth]{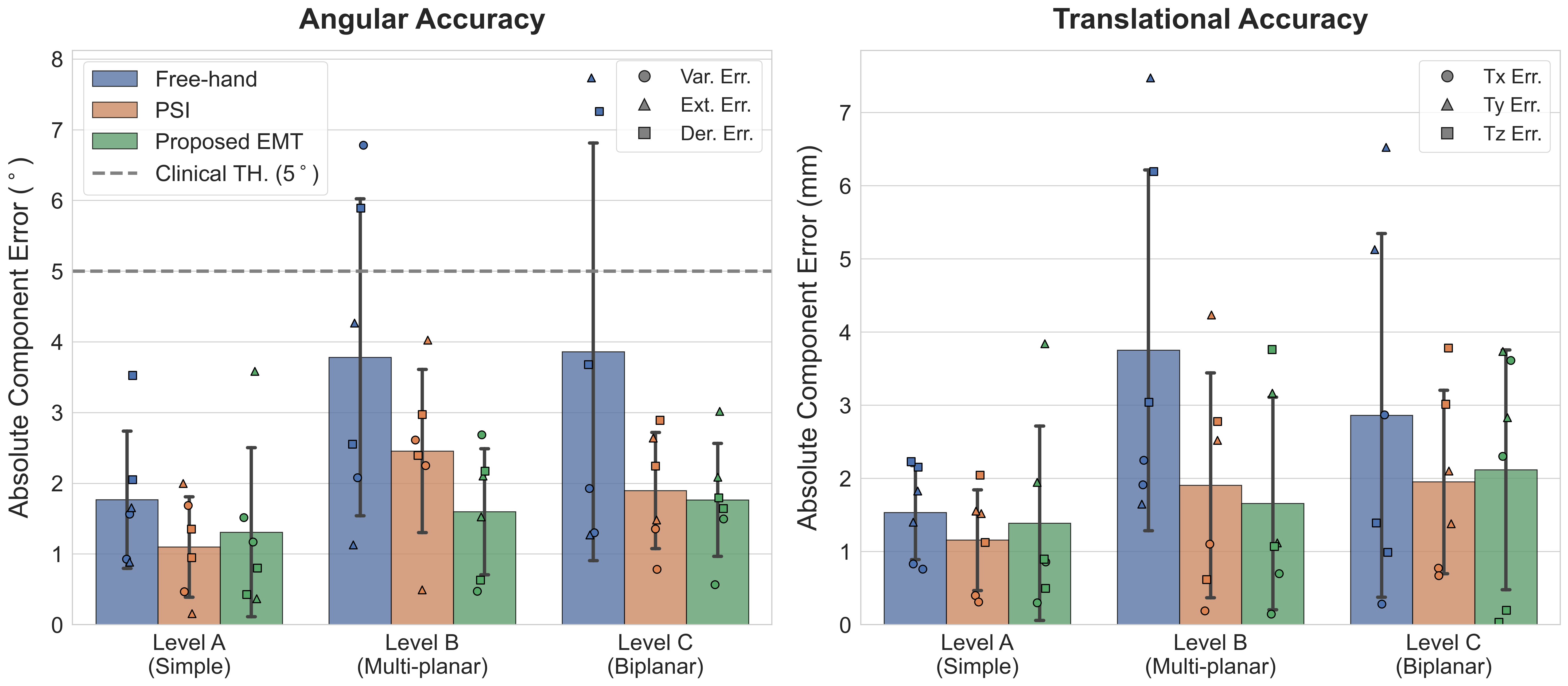}
    \caption{Angular and translational accuracy by level of difficulty. Data points represent absolute errors for each individual component (rotational and translational). The dashed line indicates the $5^\circ$ clinical threshold.}
    \label{fig:data_plot}
\end{figure}

We report the results from our comparative synthetic bone feasibility study assessing correction accuracy of our EMT system against free-hand and PSI techniques as well as an user feedback analysis.

\subsection{Phantom Feasibility Study}
\label{sec:surg_acc}

The core evaluation comprised $N=18$ osteotomies (two surgeons, three difficulty levels, three methods) in a repeated-measures design. ~\cref{fig:data_plot}, ~\cref{tab:total_euclidean_errors} and ~\cref{tab:error_components_tall} summarize the results. 

\subsubsection{Angular Accuracy}
\noindent\textbf{Comparison to Free-hand (Superiority)}
Angular accuracy of the achieved corrections was assessed and specifically compared to the free-hand technique for superiority.

\begin{itemize}
    \item \textbf{Total Angular Error}: The proposed EMT system achieved a mean Euclidean angular error of $3.05^\circ \pm 0.75^\circ$, compared to $6.32^\circ \pm 2.36^\circ$ for free-hand (mean difference: $3.27^\circ$, 95\% CI: $[0.77, 5.77]$). This reduction was statistically significant (Wilcoxon $W=20$, $p=0.031$).
    \item \textbf{Derotation:} A significant improvement was observed in the roll axis. EMT reduced the error to $1.24^\circ \pm 0.72^\circ$ compared to $4.16^\circ \pm 2.01^\circ$ for free-hand (mean difference: $2.92^\circ$, 95\% CI: $[1.39, 4.44]$; Wilcoxon $W=21$, $p=0.016$), indicating that the navigation system is particularly effective at controlling derotation alignment.
    \item \textbf{Extension:} The EMT system resulted in an error of $2.11^\circ \pm 1.13^\circ$, compared to $2.82^\circ \pm 2.71^\circ$ for free-hand. This reduction was not statistically significant (Wilcoxon $W=12$, $p=0.422$).
    \item \textbf{Varisation:} EMT resulted in a lower mean error ($1.32^\circ \pm 0.81^\circ$) than free-hand ($2.43^\circ \pm 2.17^\circ$), though this difference did not reach significance (Wilcoxon $W=16$, $p=0.156$).
\end{itemize}
The EMT system demonstrated a 52\% reduction in mean Euclidean angular error compared to the free-hand technique, alongside reduced errors across all individual angular components.

\begin{table*}[t]
\centering
\resizebox{\textwidth}{!}{

\begin{tabular}{lcccccc}
\toprule
& \multicolumn{3}{c}{\textbf{Total Angular Error} ($^\circ$)} 
& \multicolumn{3}{c}{\textbf{Total Translational Error} (mm)} \\
\cmidrule(lr){2-4} \cmidrule(lr){5-7}

\textbf{Method} & \textbf{Free-hand} & \textbf{PSI} & \textbf{EMT} 
& \textbf{Free-hand} & \textbf{PSI} & \textbf{EMT} \\
\midrule

Mean 
& $6.32 \pm 2.36$          & $3.44 \pm 1.16$          & $\mathbf{3.05 \pm 0.75}$
& $5.56 \pm 2.25$          & $\mathbf{3.36 \pm 1.18}$ & $3.59 \pm 1.44$ \\

\bottomrule
\end{tabular}
}
\caption{Comparison of mean Euclidean translation and angular errors across all difficulty levels. Values are mean $\pm$ SD.}
\label{tab:total_euclidean_errors}
\end{table*}

\noindent\textbf{Comparison to PSI (equivalence)}
To assess whether the navigation workflow could replicate the accuracy of patient-specific guides (acting as the positive control), we benchmarked the proposed EMT solution against PSI using a TOST with an equivalence margin of $\pm 2^\circ$. This analysis demonstrated statistical equivalence across all rotational components, specifically for total angular error (TOST $p=0.031$; 90\% CI: $[-0.62, 1.40]^\circ$), extension ($p=0.031$; $[-1.55, 0.92]^\circ$), varisation ($p=0.016$; $[-0.58, 0.99]^\circ$), and derotation ($p=0.031$; $[0.21, 1.57]^\circ$). In all cases, the 90\% confidence interval for the paired difference fell entirely within the $\pm 2^\circ$ equivalence bounds. These results confirm that the proposed EMT system provides angular accuracy that is statistically equivalent to PSI under these idealized conditions across all rotational degrees of freedom within a $2^\circ$ clinical tolerance.

\begin{table}[t] 
\centering
\small
\setlength{\tabcolsep}{6pt} 
\begin{tabular}{llccc}
\toprule
\textbf{Difficulty} & \textbf{Method} & \textbf{Derotation ($^\circ$)} & \textbf{Varisation ($^\circ$)} & \textbf{Extension ($^\circ$)} \\
\midrule

\textbf{Level A} 
 & Free-hand & $2.79 \pm 1.04$ & $1.24 \pm 0.45$ & $1.27 \pm 0.54$ \\
 & PSI       & $1.15 \pm 0.29$ & $\mathbf{1.08 \pm 0.86}$ & $\mathbf{1.08 \pm 1.30}$ \\
 & EMT       & $\mathbf{0.61 \pm 0.26}$ & $1.34 \pm 0.24$ & $1.97 \pm 2.27$ \\
\midrule

\textbf{Level B} 
 & Free-hand & $4.22 \pm 2.36$ & $4.43 \pm 3.32$ & $2.70 \pm 2.22$ \\
 & PSI       & $2.68 \pm 0.41$ & $2.43 \pm 0.26$ & $2.26 \pm 2.50$ \\
 & EMT       & $\mathbf{1.40 \pm 1.09}$ & $\mathbf{1.58 \pm 1.56}$ & $\mathbf{1.81 \pm 0.41}$ \\
\midrule

\textbf{Level C} 
 & Free-hand & $5.47 \pm 2.53$ & $1.61 \pm 0.44$ & $4.50 \pm 4.57$ \\
 & PSI       & $2.57 \pm 0.46$ & $1.07 \pm 0.40$ & $\mathbf{2.06 \pm 0.82}$ \\
 & EMT       & $\mathbf{1.72 \pm 0.11}$ & $\mathbf{1.03 \pm 0.66}$ & $2.55 \pm 0.66$ \\
\midrule

\textbf{Mean} 
 & Free-hand & $4.16 \pm 2.01$ & $2.43 \pm 2.17$ & $2.82 \pm 2.71$ \\
 & PSI       & $2.13 \pm 0.82$ & $1.52 \pm 0.83$ & $\mathbf{1.80 \pm 1.43}$ \\
 & EMT       & $\mathbf{1.24 \pm 0.72}$ & $\mathbf{1.32 \pm 0.81}$ & $2.11 \pm 1.13$ \\

\bottomrule
\end{tabular}
\caption{Comparison of absolute normalized error components across difficulty levels and techniques.}
\label{tab:error_components_tall}
\end{table}

\noindent \textbf{Clinical safety \& outlier analysis}
Free-hand execution demonstrated limited reliability, with 4 out of 6 trials exceeding the $5^\circ$ clinical threshold in at least one angular component, leading to deviations of up to $7.73^\circ$. Only the 2 trials on the simplest level (A) were within bounds. Conversely, the EMT system demonstrated a maximum error of $3.58^\circ$. This reliability matches that of the PSI group (max. $4.02^\circ$), suggesting that the EMT workflow eliminated the risk of significant malalignment in this feasibility study. A further qualitative advantage observed was the ability to actively monitor the registration. While the fit of a PSI guide is static and difficult to verify once pinned, our navigation system provided continuous visual feedback. This adds a layer of safety not present in the blind PSI or free-hand techniques.

\subsubsection{Translational accuracy}
EMT reduced mean translational error relative to free-hand ($3.59$\,mm vs.\ $5.56$\,mm; mean difference $1.97$\,mm, 95\% CI: $[-0.90, 4.85]$), although this difference did not reach statistical significance (Wilcoxon $W=18$, $p=0.078$). Compared to PSI, EMT showed statistical equivalence within a $\pm 2$\,mm margin (TOST $p=0.047$; 90\% CI: $[-1.97, 1.51]$\,mm), with the mean difference of $-0.23$\,mm falling entirely within bounds.

\subsection{Workload and Usability Analysis}
\label{sec:tlx}

 \cref{tab:nasa-tlx-mean} summarizes the NASA-TLX questionaire \cite{hart1988development}. EMT demonstrated the lowest workload burden across all dimensions, with the exception of mental demand, where it remained comparable to PSI ($55.0$). Notably, EMT yielded superior self-assessed performance ($12.5$), scoring less than half that of PSI and free-hand (both $27.5$), suggesting successful workflow integration.

\begin{table}[ht]
    \centering
    
    \resizebox{\columnwidth}{!}{%
        \begin{tabular}{l c c c c c c}
            \toprule
            \textbf{Method} & \textbf{Mental Demand} & \textbf{Physical Demand} & \textbf{Temporal Demand} & \textbf{Performance} & \textbf{Effort} & \textbf{Frustration} \\
            \midrule
            EMT       & 55.0 & 30.0 & 35.0 & 12.5 & 50.0 & 20.0 \\
            PSI       & 55.0 & 40.0 & 50.0 & 27.5 & 60.0 & 25.0 \\
            Free-hand & 60.0 & 35.0 & 45.0 & 27.5 & 60.0 & 30.0 \\
            \bottomrule 
        \end{tabular}%
    }
    \caption{Mean NASA-TLX scores by interaction method.}
    \label{tab:nasa-tlx-mean}
\end{table}

\section{Discussion}

\noindent\textbf{Limitations and future work} This study was designed as a preclinical feasibility study utilizing synthetic bone models. While these models provide a standardized geometric baseline for system testing, they do not replicate the properties of soft tissues or dynamic muscular forces. Consequently, while standard K-wire fixation proved sufficient for this synthetic substrate, clinical translation will likely necessitate threaded pins to ensure tracker rigidity against soft-tissue tension. Furthermore, the electromagnetic environment was controlled; although instrument interference was characterized, a clinical OR introduces additional field distortions (e.g., surgical tables) that require additional validation in a realistic setting. Finally, as the primary objective was to validate technical accuracy and system integration, the sample size ($N=18$) was scoped for a pilot comparison and surgical duration was not quantified. Subsequent ex-vivo studies will aim to power the statistical analysis for workflow efficiency and operating time.

\noindent\textbf{Conclusion:} We have presented a comprehensive navigation framework that successfully integrates automatic X-ray-to-CT registration with electromagnetic tracking. The system overcomes the traditional hurdles of CT-to-EMT registration, requiring only two fluoroscopic views to establish a reliable bone reference frame. Our validation on synthetic bones confirms that this approach ensures clinically accurate rotational alignment and robustly supports complex multi-planar osteotomies. By combining the flexibility of EMT with the accuracy of preoperative planning, this system offers a low-radiation workflow with accuracy comparable to PSI in a preclinical setting.

\section*{Declarations}
This work has been supported by the OR-X, a Swiss national research infrastructure for translational surgery, and associated funding by the University of Zurich and University Hospital Balgrist. This work is based on experiments performed at the Swiss Center for Musculoskeletal Imaging, SCMI, Balgrist Campus AG, Zürich.

\noindent\textbf{Ethical Approval and Consent} Not applicable (synthetic bone study).

\noindent\textbf{Funding} This work has been funded by the Wonderland Foundation.

\noindent \textbf{Competing Interests} The authors have no relevant financial or non-financial interests to disclose.

\bibliography{sn-bibliography}

@article{dreher2012long,
  title     = {Long-term outcome of femoral derotation osteotomy in children with spastic diplegia},
  author    = {Dreher, Thomas and Wolf, Sebastian I. and Heitzmann, Daniel and Swartman, Benedikt and Schuster, Wolfram and Gantz, Simone and Hagmann, S{\'e}bastien and D{\"o}derlein, Leonhard and Braatz, Frank},
  journal   = {Gait \& Posture},
  volume    = {36},
  number    = {3},
  pages     = {467--470},
  year      = {2012},
  publisher = {Elsevier},
  doi       = {10.1016/j.gaitpost.2012.04.017}
}

@article{buly2018femoral,
  author  = {Buly, R. L. and Sosa, B. R. and Poultsides, L. A. and Caldwell, E. and Rozbruch, S. R.},
  title   = {Femoral Derotation Osteotomy in Adults for Version Abnormalities},
  journal = {Journal of the American Academy of Orthopaedic Surgeons},
  year    = {2018},
  volume  = {26},
  number  = {19},
  doi     = {10.5435/JAAOS-D-17-00623}
}

@article{sharma2001role,
  title     = {The role of knee alignment in disease progression and functional decline in knee osteoarthritis},
  author    = {Sharma, Leena and Song, Jing and Felson, David T. and Cahue, September and Shamiyeh, Eli and Dunlop, Dorothy D.},
  journal   = {JAMA},
  volume    = {286},
  number    = {2},
  pages     = {188--195},
  year      = {2001},
  publisher = {American Medical Association},
  doi       = {10.1001/jama.286.2.188}
}

@article{kriechling2025radiation,
  author  = {Kriechling, Philipp and Meichtry, Fabian Luca and Smolle, Maria Anna and Preiss, Helga and Zellner, Michael and Kaeser, Yvonne and Farshad-Amacker, Nadja A. and Dreher, Thomas},
  title   = {Analysis of intraoperative radiation exposure in paediatric orthopaedic and trauma patients: A comparative cohort study},
  journal = {BMJ Paediatrics Open},
  year    = {2025},
  volume  = {9},
  number  = {1},
  doi     = {10.1136/bmjpo-2025-003666}
}

@article{Shi2019,
  author  = {Shi, JianHui and Lv, Wei and Wang, Yan and Ma, Ben and Cui, Wei and Liu, ZhenZhong and Han, KeCheng},
  title   = {Three dimensional patient-specific printed cutting guides for closing-wedge distal femoral osteotomy},
  journal = {International Orthopaedics},
  year    = {2019},
  volume  = {43},
  number  = {3},
  pages   = {619--624},
  doi     = {10.1007/s00264-018-4043-3}
}

@article{khela2025psi_review,
  title   = {Indications, Technique, and Outcomes of Patient Specific Instrumentation for Osteotomy About the Knee},
  author  = {Khela, Harmon S. and Khela, Monty S. and Sriram, Varun and Schroeder, Grant G. and Hollyer, Ian and Sherman, Seth L.},
  journal = {Current Reviews in Musculoskeletal Medicine},
  year    = {2025},
  volume  = {18},
  pages   = {547--557},
  doi     = {10.1007/s12178-025-09987-2}
}

@article{grillo20253d,
  author    = {Grillo, Giovanni and Coelho, Alexandre and Pelfort, Xavier and Fillat-Gom{\`a}, Ferran and Verdaguer Figuerola, Arnau and Gil-Gonzalez, Sergi and Pe{\~n}alver, Juan Manuel and Yela-Verd{\'u}, Christian},
  title     = {3D-printed patient-specific instrumentation and the freehand technique in high-tibial osteotomy: A prospective cohort-comparative study in an outpatient setting},
  journal   = {Journal of Experimental Orthopaedics},
  volume    = {12},
  number    = {1},
  year      = {2025},
  doi       = {10.1002/jeo2.70088}
}

@article{casari2021ar,
  author  = {Casari, Fabio A. and Navab, Nassir and Hruby, Laura A. and Kriechling, Philipp and Nakamura, Ricardo and Tori, Romero and de Lourdes Dos Santos Nunes, F{\'a}tima and Queiroz, Marcelo C. and F{\"u}rnstahl, Philipp and Farshad, Mazda},
  title   = {Augmented Reality in Orthopedic Surgery Is Emerging from Proof of Concept Towards Clinical Studies: a Literature Review Explaining the Technology and Current State of the Art},
  journal = {Current Reviews in Musculoskeletal Medicine},
  year    = {2021},
  volume  = {14},
  number  = {2},
  pages   = {192--203},
  doi     = {10.1007/s12178-021-09699-3}
}

@article{song2016computer,
  author    = {Song, Sang Jun and Bae, Dae Kyung},
  title     = {Computer-Assisted Navigation in High Tibial Osteotomy},
  journal   = {Clinics in Orthopedic Surgery},
  volume    = {8},
  number    = {4},
  pages     = {349--357},
  year      = {2016},
  doi       = {10.4055/cios.2016.8.4.349}
}

@misc{stryker_q_guidance,
  author = {{Stryker Corporation}},
  title  = {{Ortho Q Guidance System}},
  year   = {2022},
  note   = {Version 2.0.1}
}

@article{Sorriento2020,
  author  = {Sorriento, Angela and Porfido, Maria Bianca and Mazzoleni, Stefano and Calvosa, Giuseppe and Tenucci, Miria and Ciuti, Gastone and Dario, Paolo},
  title   = {Optical and Electromagnetic Tracking Systems for Biomedical Applications: A Critical Review on Potentialities and Limitations},
  journal = {IEEE Reviews in Biomedical Engineering},
  volume  = {13},
  pages   = {212--232},
  year    = {2020},
  doi     = {10.1109/RBME.2019.2939091}
}

@manual{MedtronicStealthS8,
  title  = {StealthStation™ S8 Navigation Platform System Manual},
  author = {{Medtronic Navigation, Inc.}},
  year   = {2017},
  note   = {Version 2.0.1}
}

@article{geisbusch2025multiplanar,
  author  = {Geisb{\"u}sch, Andreas and Gramer, Carina and Dreher, Thomas and Hagen, Niclas and Hagmann, S{\'e}bastien and Renkawitz, Tobias and G{\"o}tze, Marco},
  title   = {Electromagnetic bone segment tracking in multiplanar osteotomies---A saw bone study},
  journal = {Journal of Orthopaedic Research},
  year    = {2025},
  volume  = {43},
  number  = {2},
  pages   = {362--369},
  doi     = {10.1002/jor.26000}
}

@article{liu_2020,
  author  = {Liu, He and Baena, Ferdinando Rodriguez Y},
  title   = {Automatic Markerless Registration and Tracking of the Bone for Computer-Assisted Orthopaedic Surgery}, 
  journal = {IEEE Access}, 
  year    = {2020},
  volume  = {8},
  pages   = {42010-42020},
  doi     = {10.1109/ACCESS.2020.2977072}
}

@inproceedings{Gopalakrishnan_2024,
  title     = {Intraoperative 2D/3D Image Registration via Differentiable X-Ray Rendering},
  author    = {Gopalakrishnan, Vivek and Dey, Neel and Golland, Polina},
  booktitle={CVPR}, 
  year      = {2024},
  pages     = {11662--11672},
  doi       = {10.1109/CVPR52733.2024.01108}
}

@article{Flepp2025,
  author  = {Flepp, Roman and Nissen, Leon and Sigrist, Bastian and Nieuwland, Arend and Cavalcanti, Nicola and F{\"u}rnstahl, Philipp and Dreher, Thomas and Calvet, Lilian},
  title   = {Automatic multi-view {X-ray/CT} registration using bone substructure contours},
  journal = {International Journal of Computer Assisted Radiology and Surgery},
  year    = {2025},
  volume  = {20},
  number  = {7},
  pages   = {1401--1408},
  doi     = {10.1007/s11548-025-03391-4}
}

@misc{imfusion_software,
  author       = {{ImFusion GmbH}},
  title        = {ImFusion SDK and Suite},
  howpublished = {\url{https://www.imfusion.com}},
  note         = {Accessed: 2025-12-29}
}

@incollection{hart1988development,
  author    = {Hart, Sandra G. and Staveland, Lowell E.},
  title     = {Development of {NASA-TLX} ({Task Load Index}): Results of Empirical and Theoretical Research},
  booktitle = {Human Mental Workload},
  editor    = {Hancock, Peter A. and Meshkati, Najmedin},
  publisher = {North-Holland},
  address   = {Amsterdam},
  pages     = {139--183},
  year      = {1988},
  isbn      = {978-0444703880}
}

@article{Jud_2020,
author = {Jud, Lukas and Vlachopoulos, Lazaros and Beeler, Silvan and Tondelli, Timo and Fürnstahl, Philipp and Fucentese, Sandro},
year = {2020},
title = {Accuracy of three dimensional-planned patient-specific instrumentation in femoral and tibial rotational osteotomy for patellofemoral instability},
volume = {44},
journal = {International Orthopaedics},
doi = {10.1007/s00264-020-04496-y}
}

@article{Neopoulos2025,
author = {Georgios Neopoulos and Lukas Jud and Lazaros Vlachopoulos and Sandro F. Fucentese},
title ={Combined Correction of Coronal and Rotational Deformities of the Femur With Distal Femoral Osteotomy Using Patient-Specific Instrumentation},
journal = {The American Journal of Sports Medicine},
volume = {53},
number = {4},
pages = {848-854},
year = {2025},
doi = {10.1177/03635465251314868},
}
\end{document}